\documentclass[10pt,twocolumn,letterpaper]{article}

\usepackage[pagenumbers]{cvpr} 

\usepackage{graphicx}
\usepackage{amsmath}
\usepackage{amssymb}
\usepackage{booktabs}

\usepackage[pagebackref,breaklinks,colorlinks]{hyperref}

\usepackage[capitalize]{cleveref}
\crefname{section}{Sec.}{Secs.}
\Crefname{section}{Section}{Sections}
\Crefname{table}{Table}{Tables}
\crefname{table}{Tab.}{Tabs.}

\def\cvprPaperID{*****} 
\def\confName{CVPR}
\def\confYear{2023}

\begin{document}

\title{SFGANS: Self-supervised Future Generator for human ActioN Segmentation}

\author{Or Berman\\
Technion\\
{\tt\small or-berman@campus.technion.ac.il}
\and
Adam Goldbraik\\
Technion\\
{\tt\small goldb.adam@gmail.com}
\and
Shlomi Laufer\\
Technion\\
{\tt\small laufer@technion.ac.il}
}
\maketitle

\begin{abstract}
   

The ability to locate and classify action segments in long untrimmed video is of particular interest to many applications such as autonomous cars, robotics and healthcare applications. Today, the most popular pipeline for action segmentation is composed of encoding the frames into feature vectors, which are then processed by a temporal model for segmentation. In this paper we present a self-supervised method that comes in the middle of the standard pipeline and generated refined representations of the original feature vectors. Experiments show that this method improves the performance of existing models on different sub-tasks of action segmentation, even without additional hyper parameter tuning.
\end{abstract}

\section{Introduction}
\label{sec:intro}
Human action segmentation is a crucial and fundamental task for many applications, including surveillance, robotics, security, surgical applications, and autonomous cars. The task of action segmentation is considered to be time consuming as it requires labeling of each video frame. Moreover, several applications, such as in healthcare, require online performance. As a result, different forms of action segmentation have been developed, such as online action segmentation \cite{goldbraikh2022bounded}, and forms of weekly supervised action segmentation \cite{li2021temporal}. In contrast to action segmentation, classifying short trimmed videos with a single label, also known as 'action recognition', has been considered a more simple and straightforward task \cite{wang2021temporal, wang2020boundary}.

Over the years different methods were implemented to tackle this task. Early works applied the sliding window approach\cite{karaman2014fast, rohrbach2012database}, while others used different combinations of convolutional  neural networks (CNNs) and recurrent neural networks (RNNs) \cite{donahue2015long, vinyals2015show}.\\
Today, the main approach consist of two stages. First, a pre-trained network is used to encode frames into feature vectors. Then, a temporal model predicts action labels for each frame based on those features. \cite{lea2017temporal, farha2019ms, li2020ms, chinayi_ASformer, wang2021temporal, wang2020boundary}.

In the world of self-supervised learning, there are two main approaches. One approach trains the model on a self-supervised task to initialize the weights and provide the model with domain knowledge. The other employ a self supervised task on a model in order to receive a more informative data representation. Self supervision has been proven to boost model performance and creates better data representations \cite{doersch2015unsupervised, gidaris2018unsupervised, xu2019self, zbontar2021barlow}.

\begin{figure}
    \centering
    \includegraphics[width=\linewidth]{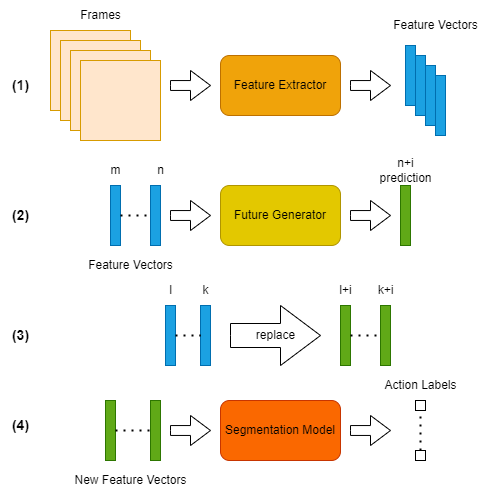}
    \caption{Full paper pipeline. (1) Frames are encoded into feature vectors using a feature extractor. (2) A prediction of the (n+i)-th vector is generated using a sequence of feature vectors. In this paper we implemented for i values of 1, 4, and 10. (3) Replacing the n-th feature vector with prediction n+i. Phases 2 and 3 are repeated for each vector. (4) The new predicted features replaces the original ones,  and sent to the segmentation model.}
    \label{fig:pipeline}
\end{figure}

In contrast to most self-supervised approaches that are executed on the backbone model or the input data, this paper presents a novel approach for the human action segmentation pipeline. Instead of using the feature vectors produced by the backbone, we trained a model that predicts the future features in the sequence, based on the original features. Then, we replaced the feature vectors with the future predictions, as explained in figure \ref{fig:pipeline}. 

Our contributions are as follows:
\begin{itemize}
    \item An additional self-supervised stage to the current standard pipeline, that boost performance of existing models for human action segmentation.
    \item The method proved to boost performance on sub-tasks of human action segmentation: online action segmentation and timestamp supervision action segmentation.    
    \item This method boost performance on existing models without the need for hyperparameter tuning. 
    \item The method improves performance metrics on different datasets from different domains.
    \item The method proved to improve results with different backbones/feature extractors.

\end{itemize}

\section{Related Work}
\textbf{Action segmentation.} Early works attempted to identify actions of different lengths by applying the sliding window technique. For example, Rohrbach \etal. \cite{rohrbach2012database} used SVMs with sliding window to find different actions on their new dataset at the time. Later works applied different deep learning techniques to tackle the task. Donahue \etal. \cite{donahue2015long} used combinations of 2D CNNs with RNNs in order to predict actions. Singh \etal. \cite{singh2016multi} also used recurrent networks, but instead combined them with multi-stream 2D CNN, combining RGB data and optical flow data. Lea \etal \cite{lea2017temporal} presented 'TCN' - temporal convolutional network that applied temporal convolutions with dilation on feature vectors extracted by a pre-trained CNN. In recent years it became popular to use feature vectors as inputs for the task of action segmentation. Moreover, TCNs also became commonly used in models for such tasks. Farha \etal. \cite{farha2019ms} for example, presented 'MS-TCN', which by itself became a very popular backbone, and was based on Lea's TCN. MS-TCN operates on feature vectors extracted from I3D \cite{carreira2017quo}. Similar to MS-TCN, most current action recognition methods consist of 2 major components: a pre-trained model to extract feature vectors and a segmentation model to classify each vector into an action. Wang \etal. \cite{wang2020boundary} used the I3D features as an input and procceseed them using stage cascades combined with MS-TCN. Yi \etal.  \cite{chinayi_ASformer} also used the I3D features as the input to a modefied transformer.

\textbf{Self supervision.} These tasks increase models' knowledge and performance using unlabeled data. It is done by defining a task in which the data is also used as labels. This method provides models with more domain knowledge or make the data more informative. Self-supervision has a large number of applications, ranging from robotics to image understanding \cite{kolesnikov2019revisiting}.

Doersch \etal. \cite{doersch2015unsupervised} proposed a method that divides the images to patches and predicted their correct location in the image. Gidaris \etal. \cite{gidaris2018unsupervised} presented a method of predicting the rotation angle of a rotated image out of four rotation possibilities. Xu \etal. \cite{xu2019self} learned spatio-temporal information by predicting the correct order of shuffled video clips. Zbontar \etal. \cite{zbontar2021barlow} presented Barlow twins as a method of learning embeddings by training models to output similar vectors for distorted versions of the same image.

 Another popular task for creating better embeddings or reinforce model's knowledge is predicting the future. For example Finn \etal. \cite{finn2016unsupervised} developed an action-conditioned video prediction model that explicitly models pixel motion, by predicting a distribution over pixel motion from previous frames. Srivastava \etal. \cite{srivastava2015unsupervised} learned representations of videos using LSTMs that encode videos and then tried to reconstruct them and predicted future sequences.

\textbf{Future prediction.} Future prediction is an extensive field of research. While some methods like TOS\_AF\_TSC \cite{ke2019time} and AVT \cite{girdhar2021anticipative} are trying to predict the next action in a video sequence, many others are trying to predict sequences of future frames. A main reason for this is the rise of interest in the filed of autonomous cars, and the understanding that the knowledge a model gains from predicting frames is useful for many tasks. 
Today, the main tools for next frame prediction tasks are GANs (Generative Adversarial Models) and VAEs (Variational Auto Encoders). For example, DMGAN \cite{liang2017dual} uses 2 generators to predict the next frame and the next optical flow, and uses a combination of adversarial loss and VAE loss. Vondrick \etal.\cite{vondrick2016generating} created a video generator and converted it into a next frame prediction model. Mathieu \etal.\cite{mathieu2016deep} used nested GANs to generate next frames on different resolutions. Kwon \etal. \cite{kwon2019predicting} presented a GAN with a retrospective training procedure. 
In contrast, Villegas \etal.\cite{villegas2017learning} and Hosseini \etal.\cite{hosseini2019inception} preferred to use an LSTM based approach. Villegase used domain knowledge and predicted the next frames by generating human pose estimations, and warp them with a past frame to change the state of the human in the image. The use of poses instead of frames enabled them to work with feature vectors, which are much smaller then frames, creating a more efficient model with relatively small number of parameters. Hosseini created an inception LSTM, based on convolutional LSTM\cite{lotter2017deep}, to predict the next frames. In the medical field, Gao \etal.\cite{gao2021future} used LSTMs encoders-decoders with VAE settings to predict motion distribution and content distribution on the Jigsaws dataset\cite{gao2014jhu}. 

\section{Method}
\subsection{Future Prediction}\label{Future Prediction}
We define feature vectors sequence of length $n$ as:
\begin{equation}\label{eqn:sequence}
    V_{m:m+n-1} = \{v_m, v_{m+1}, ..., v_{m+n-1}\}, n > 0
\end{equation}
where $v_{i}\in \mathbb{R}^k$ is the feature vector for the $i$-th frame in a video and $k$ is the dimension of the feature vector. Using these features as inputs, the model's goal is to generate the vector $\hat{v}_{m+n}$ that is as similar as possible to the real next feature vector in the sequence, $v_{m+n}$. The metrics for calculating similarity are defined in section \ref{Metrics}.

In a $\ell$-future prediction setting the goal is to generate the sequence
\begin{equation}\label{eq:l-future}
    \hat{V}_{m+n:m+n+l-1} = \{\hat{v}_{m+n}, \hat{v}_{m+n+1},..., \hat{v}_{m+n+l-1}\}
\end{equation}
that is as similar as possible to the real sequence, $V_{m+n:m+n+l-1}$. In most cases, including this paper, this prediction is done iteratively. In each iteration the generated future vector is added to the end of the input sequence, and the first vector of this sequence is dropped. The modified sequence is the input to the model in the next iteration. For example, in order to generate $\hat{v}_{m+n+1}$ the model will generate $\hat{v}_{m+n}$ using the sequence in \ref{eqn:sequence}, and than generate $\hat{v}_{n+m+1}$ using:
\begin{equation}
    \{v_{m+1}, v_{m+2},..., v_{m+n-1}, \hat{v}_{m+n}\}
\end{equation}

\subsection{Model}\label{model}
\begin{figure*}
    \centering
    \includegraphics[width=\linewidth]{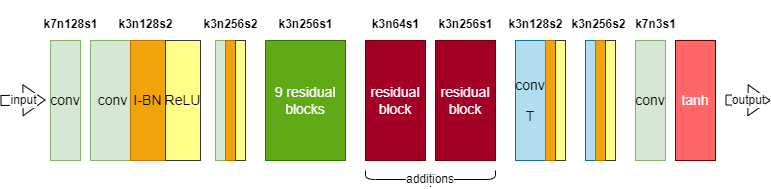}
    \caption{The generator architecture. It is strongly based on the original generator architecture from \cite{kwon2019predicting} with additions marked in red. In the figure, I-BN is instance batch norm, and k, n, and s denote the kernel size, channels, and stride respectively. A More detailed description of the architecture, including the architectures of the residual blocks and the discriminators' are found in \cite{kwon2019predicting}. The notations are similar for convenience.}
    \label{fig:generator}
\end{figure*}
In order to predict future feature vectors we adopted the retrospective cycle GAN \cite{kwon2019predicting} framework, which originally used to predict future frames of a video sequence. The framework consists of a generator, that predicts future feature vectors, and 2 discriminators. One discriminator predicts if a feature vector is real or generated and the other predicts if a sequence of feature vectors contains only real feature vectors, or a mix of real and generated feature vectors. The discriminators' architectures are similar to the ones presented in \cite{kwon2019predicting} with the modification of using 1D convolutions instead of 2D convolutions.

The generator was designed based on the original generator architecture with 1D convolutions and an addition of 2 residual blocks in the center of the model. The role of the additional blocks is to drastically reduce the number of channels of the input from 256 to 64 and then immediately back to 256. It can be seen as a division of the generator into an encoder and a decoder, and was done in order to help the model generalize and extract the most significant information from the past feature vectors. More details can be found in figure \ref{fig:generator}. A convolutional architecture was chosen since past research indicates that CNN are superior to recurrent networks, such as GRUs and LSTM, in sequence modeling tasks \cite{bai2018empirical}. These architectures are also common in time series forecasting \cite{lim2021time}. 

The inputs to the model are $n$ consecutive feature vectors and the output is the prediction of the next feature vector. In order to predict farther ahead into the future, one should only drop the first vectors in the sequence, and append the predictions iteratively until getting the wanted prediction (as explained in \ref{Future Prediction}). The model is also trained to predict past features and it can be done by reversing the sequence order.

\subsection{Training Procedure}
A sequence such as in \ref{eqn:sequence} is used in order to generate $\hat{v}_{m+n}$. The reversed sequence, $V_{m+n:m+1}$, is used to generate $\hat{v}_{m}$. The sequence presented in \ref{eqn:sequence} is then reconstructed using $\hat{v}_{m}$ instead of $v_m$ and then the generator uses it to predict another future vector $\hat{\hat{v}}_{m+n}$. Similar process happens to create $\hat{\hat{v}}_{m}$.
Moreover, the original prediction $\hat{v}_{m+n}$ is then used iteratively as explained in \ref{Future Prediction} in order to predict a total of $l$ future vectors. This was added to enable and enhance more accurate long term future prediction \cite{villegas2017learning}. We set $l$ to 10 in all our experiments.

\subsection{Objective Function}
The loss function is defined as follows:
\begin{equation}
    L = L_{cyc} + \lambda_{seq} \cdot L_{seq}
\end{equation}
$L_{cyc}$ is the original loss function presented in \cite{kwon2019predicting}. It consists of 4 elements: 2 adversarial losses (one for the frame discriminator and one for the sequence discriminator), MSE loss between the predicted and original vectors and MSE loss between the Laplacian of Gaussian (LoG) of the original and predicted vector.

$L_{seq}$ is the MSE between the predicted $l$-length sequence and the original $l$-length sequence. $\lambda_{seq}$ is a weight for $L_{seq}$. $\lambda_{seq}$ is 0.003 in all our experiments.

\subsection{Metrics}\label{Metrics}
We used 3 popular metrics to compute similarity between the predicted feature vector and the original one:
\textbf{MSE} is the mean over the errors of the prediction relative to the original vectors. It computes as follows:
\begin{equation}
    MSE(v, \hat{v}) = \sum_{i=1}^{k} |v^{(i)} - \hat{v}^{(i)}|^2
\end{equation}
where $v^{(i)}$ is the value of vector $v$ at index $i$.

\textbf{PSNR} \cite{huynh2008scope} or peak signal-to-noise ratio is the ratio between the maximum possible power of a signal and the power of corrupting noise that affects its quality. To estimate the PSNR of a signal, it is necessary to compare that signal to an ideal clean signal with the maximum possible power. As accepted in frame prediction, we compute the PSNR between the prediction and the original vectors as follows:
\begin{equation}
    PSNR(v, \hat{v}) = 10*log_{10}(\frac{MAX_{I}^2}{\sqrt{MSE(v, \hat{v})}})
\end{equation}
Where $MAX_{I}$ is the maximal possible value.
To alignment with the filed of next frame prediction, the features were normalized to range of 0-255 before calculations, hence $MAX_{I}$ is 255.

\textbf{SSIM} \cite{wang2004image} stands for structural similarity. In image processing, structural information refers to the inter-dependency between pixels, especially when they are close to one another. These dependencies carry significant information about the structure of the objects in the image. This can also be true for multi dimensional signals like feature vectors. SSIM is computed as follows:
\begin{equation}
    SSIM(v, \hat{v}) = \frac{(2\cdot\mu_{v}\cdot\mu_{\hat{v}}+c1)\cdot(2\cdot\sigma+c2)}{(\mu_{v}^2+\mu_{\hat{v}}^2+c1)\cdot(\sigma_{v}^2+\sigma_{\hat{v}}^2+c2)}
\end{equation}
Where $\sigma$ is the covariance between $v$ and $\hat{v}$ and $\mu_{v}$, $\mu_{\hat{v}}$,  $\sigma_{v}, \sigma_{\hat{v}}$ are the expectations and standard deviations of $v$ and $\hat{v}$ respectively. $c1$ and $c2$ are smoothing parameters that depend on the range between values.

\section{Tasks, Models and Datasets}
We assessed the usage of the presented model as a self-supervised technique for performance boost on several tasks on several datasets.
This was achieved by replacing the original features with the predicted future features, as shown in \ref{fig:pipeline}.

We applied the described method on different state-of-the-art models for several sub-tasks of action segmentation and compared their results with the original reconstructed results.

The rest of the section is as follows:
For each sub-task we describe the task in hand and the models that were tested on it. Afterwards, we described the datasets and which sub-tasks each dataset was used for.

\subsection{Temporal Action Segmentation}
\textbf{Temporal Action Segmentation} aims to segment each video frame with pre-defined action labels. For example, for a video with 300 frames the neural network will output 300 labels, one for each frame. Occasionally, this task is also called 'action segmentation'.

\subsubsection{Models}
The task of temporal action segmentation was evaluated on the following models:

\textbf{MS-TCN++} \cite{li2020ms} stands for 'multi stage temporal convolutional network', is an extended variation of the original MS-TCN \cite{farha2019ms}. Although it has the lowest performance compared to other models we presented, the model architecture and relatively short runtimes made it popular as a skeleton for many future works in the field of video processing and understanding\cite{wang2021temporal, wang2020boundary, goldbraikh2022bounded, chen2020action, zhang2021ms, ma2021towards}. MS-TCN++ uses dual dilated convolutional layers to simultaneously receive information from different time spans, and then use refinement units to refine the initial prediction.

\textbf{ASFormer} \cite{chinayi_ASformer} is a model composed of a modified transformer encoder, designed do deal with long sequences, and 3 transformer decoders to refine the initial prediction, inspired by the work of \cite{li2020ms, wang2020boundary}. To the best of our knowledge ASFormer has the highest performance for this task.

\textbf{DGTRM} \cite{wang2021temporal} stands for Dilated Temporal Graph Reasoning Module. The network attempts to model temporal relations and dependencies between video frames at various time spans using graph neural networks. This module is designed as the last stage of an action segmentation model. In their paper, Wang \etal. used the dilated TCN from MS-TCN \cite{farha2019ms} as a backbone.

\subsection{Timestamp Supervision Temporal Action Segmentation}
\textbf{Timestamp Supervision Temporal Action Segmentation}  \cite{li2021temporal, moltisanti2019action} is a task similar to temporal action segmentation, with much weaker supervision. Instead of having the ground truth for each frame in the dataset, there is a single labeled frame for each action instance. The motivation is to minimize the time and financial costs of data labeling.

\subsubsection{Model}
In this work we used a model created by Li \etal. \cite{li2021temporal}. The model is a refined MS-TCN fused with stamp-to-stamp energy function and a confidence loss, in order to predict changes between actions.

\subsection{Online Action Segmentation}
\textbf{Online action segmentation} \cite{twinanda2016endonet, goldbraikh2022bounded} is of great importance when predictions must occur as fast as possible. For example, in the medical domain, where fast decisions are crucial. The label prediction of a frame is based only on data from previous and current frames. In regular action segmentation, a frame can be classified based on both past and future information.

\subsubsection{Model}
MS-TCN++ with different dilation and padding in order to create an online model. Based on the work of Goldbraikh \etal. \cite{goldbraikh2022bounded}. 

\subsection{Datasets}
We evaluated the temporal action segmentation sub-tasks on 3 major datasets from the domain of cooking/food making, and one dataset from the medical domain. Each model was tested on the original paper datasets. We reconstructed the original features' results for each model and dataset. For the 3 food making datasets, the features are the I3D \cite{carreira2017quo} 2048 dimension feature vectors proposed in \cite{farha2019ms}. For the medical dataset, the features were extracted from Efficient-Net as presented in \cite{goldbraikh2022bounded}. 

\textbf{Breakfast} \cite{kuehne2014language} is the largest dataset we used. It contains over 77 hours of videos with a total of 1712 videos recorded in 18 different kitchens. It shows 52 participants making different breakfasts from 3-5 points of view and an overall of 48 different actions.
Each video contains 6 action instances on average. For evaluation we used the standard 4 splits, performing cross validation.
The dataset was used for action segmentation and timestamp supervision action segmentation.

\textbf{50salads} \cite{stein2013combining}  contains over 4.5 hours of videos documenting people making different salads. It has a total of 50 videos of 25 actors, where each actor made 2 different salads. Each video is 6.4 minutes long on average and contains about 20 action instances. There are 17 action classes. For evaluation we used the standard 5 splits, performing cross validation.
The dataset was used for action segmentation and timestamp supervision action segmentation.

\textbf{GTEA} \cite{fathi2011learning} also known as Georgia Tech Egocentric Activities. It is the smallest dataset in this paper, containing 28 videos of 4 actors, each perform 7 different tasks of making food. The videos were recorded using a camera mounted on the actor's head. There is a total of 11 different actions classes in the dataset, and each video contains 20 action instances on average.
The 4 standard splits were use. Each split has one actor as the test set and the others as train.
The dataset was used for action segmentation and timestamp supervision action segmentation.

\textbf{VTS} \cite{goldbraikh2022video} stands for Variable Tissue Simulator. It is a medical dataset with videos of people suturing 2 different types of tissues. We used this dataset for the task of online action segmentation. There are 24 participants in the dataset, where each performed the task twice on each tissue. The dataset contains 96 videos with an average of 2-6 minutes for each video. There are 6 action classes. For evaluation we used the standard 5 splits. Each split contains a train, validation and test set. We didn't use the test set at all during the training of the future prediction model. For this dataset we used pre-extracted Efficient-Net features from \cite{goldbraikh2022bounded}.
The dataset was used for action segmentation and online action segmentation.

\section{Experiments and Results}
\subsection{Future Prediction}
For each dataset we trained the future generator from \ref{model}. We trained a model for each split of the dataset. The models were trained at the same time, and we stop training once all models had converged. Convergence was measured by the metrics mentioned in section \ref{Metrics}. Then, for each split, we choose a future predictor using the following metric:

First, we normalized each metric over the epochs using min-max normalization, so all metrics will be on the same scale. Since lower MSE means more similarity, we reversed the MSE for each epoch:
\begin{equation}
    \hat{MSE} = 1 - MSE
\end{equation}
Afterwards, we chose the 25 epochs with the highest test metrics' mean. From those 25 epochs we chose the epoch with test mean that is the closest to it's train mean. This was done to ensure that the test's features distribution will be as similar as possible to the train's features distribution. We used the model from this epoch as a future generator for this split.

We trained the model using automatic mixed precision \cite{micikevicius2017mixed} with the Adam optimizer, $lr=0.00003$, $\beta_1=0.5$, and $\beta_2=0.999$. The model input sequence length was 20. 

\subsection{Human Action Segmentation}
We experimented with our new features on 3 action segmentation tasks: regular action segmentation, online action segmentation, and timestamp supervision action segmentation. We tried different sets of features that can be produced by our method and compared their results to the original results. The original results were reconstructed updated CUDA, pytorch, and python versions, using the models' published code and hyper parameters presented in their papers. In all experiments, unless specified, we used the original hyper parameters without fine-tuning. This was done to examine if one can use our new features and receive better results without any additional fine-tune. Moreover, for each task and model, the reconstruction and our experiments were done on the same machine under the same terms. 

\subsubsection{Action Segmentation}\label{sec:regular}
In this section we present the results on the task of regular action segmentation. We tested 4 models on 4 datasets. We tested ASFormer \cite{chinayi_ASformer}, DTGRM \cite{wang2021temporal}, and MS-TCN++ \cite{li2020ms} on 50salads \cite{stein2013combining}, GTEA \cite{fathi2011learning} and Breakfast \cite{kuehne2014language}. We also tested a version of MS-TCN++, created by \cite{goldbraikh2022bounded}, on VTS \cite{goldbraikh2022video}. As explained in equation \ref{eq:l-future}, we can predict more than a single future vector. Therefore, we have experimented with different sets of features: \textbf{1-encoded} - prediction of the next feature vector, \textbf{4-encoded} - the prediction of the 4th next feature vector. For the VTS dataset we also generated \textbf{10-encoded} - the prediction of the 10th next feature vector. \\
The results are presented in tables \ref{tab:ASFormer results}, \ref{tab:MS-TCN++ results}, \ref{tab: DTGRM results} and \ref{tab:APAS}.

\begin{table*}[!h]
\begin{tabular}{|c|ccccc|ccccc|ccccc|}
\hline
          & \multicolumn{5}{c|}{50salads}                                                                                                                                     & \multicolumn{5}{c|}{GTEA}                                                                                                                                         & \multicolumn{5}{c|}{Breakfast}                                                                                                                                    \\ \hline
          & \multicolumn{1}{c|}{Acc}           & \multicolumn{1}{c|}{Edit}          & \multicolumn{3}{c|}{F1@\{10, 25, 50)}                                                   & \multicolumn{1}{c|}{Acc}           & \multicolumn{1}{c|}{Edit}          & \multicolumn{3}{c|}{F1@\{10, 25, 50\}}                                                  & \multicolumn{1}{c|}{Acc}           & \multicolumn{1}{c|}{Edit}          & \multicolumn{3}{c|}{F1@\{10, 25, 50\}}                                                  \\ \hline
Original  & \multicolumn{1}{c|}{85.8}          & \multicolumn{1}{c|}{76.3}          & \multicolumn{1}{c|}{83.0}          & \multicolumn{1}{c|}{81.6}          & 74.6          & \multicolumn{1}{c|}{\textbf{80.1}} & \multicolumn{1}{c|}{85.5}          & \multicolumn{1}{c|}{\textbf{90.7}} & \multicolumn{1}{c|}{\textbf{89.8}} & 79.4          & \multicolumn{1}{c|}{74.0}          & \multicolumn{1}{c|}{75.4}          & \multicolumn{1}{c|}{76.6}          & \multicolumn{1}{c|}{71.6}          & 58.9          \\ \hline
1-Encoded   & \multicolumn{1}{c|}{85.2}          & \multicolumn{1}{c|}{\textbf{78.7}} & \multicolumn{1}{c|}{\textbf{85.0}} & \multicolumn{1}{c|}{\textbf{83.3}} & 76.3          & \multicolumn{1}{c|}{79.5}          & \multicolumn{1}{c|}{85.9}          & \multicolumn{1}{c|}{89.8}          & \multicolumn{1}{c|}{88.6}          & 79.7          & \multicolumn{1}{c|}{\textbf{74.7}} & \multicolumn{1}{c|}{75.6}          & \multicolumn{1}{c|}{77.1}          & \multicolumn{1}{c|}{71.7}          & 59.4          \\ \hline
4-Encoded & \multicolumn{1}{c|}{\textbf{85.9}} & \multicolumn{1}{c|}{77.2}          & \multicolumn{1}{c|}{84.3}          & \multicolumn{1}{c|}{82.7}          & \textbf{76.4} & \multicolumn{1}{c|}{78.7}          & \multicolumn{1}{c|}{\textbf{86.3}} & \multicolumn{1}{c|}{90.0}          & \multicolumn{1}{c|}{88.2}          & \textbf{80.6} & \multicolumn{1}{c|}{74.5}          & \multicolumn{1}{c|}{\textbf{75.9}} & \multicolumn{1}{c|}{\textbf{77.2}} & \multicolumn{1}{c|}{\textbf{72.2}} & \textbf{59.6} \\ \hline
\end{tabular}
\caption{ASFormer results. There is an improvement in most metrics, with some improved by over 2\%.}
\label{tab:ASFormer results}
\end{table*}

\begin{table*}[!h]
\begin{tabular}{|c|ccccc|ccccc|ccccc|}
\hline
          & \multicolumn{5}{c|}{50salads}                                                                                                                                     & \multicolumn{5}{c|}{GTEA}                                                                                                                                         & \multicolumn{5}{c|}{Breakfast}                                                                                                                                    \\ \hline
          & \multicolumn{1}{c|}{Acc}           & \multicolumn{1}{c|}{Edit}          & \multicolumn{3}{c|}{F1@\{10, 25, 50)}                                                   & \multicolumn{1}{c|}{Acc}           & \multicolumn{1}{c|}{Edit}          & \multicolumn{3}{c|}{F1@\{10, 25, 50\}}                                                  & \multicolumn{1}{c|}{Acc}           & \multicolumn{1}{c|}{Edit}          & \multicolumn{3}{c|}{F1@\{10, 25, 50\}}                                                  \\ \hline
Original  & \multicolumn{1}{c|}{82.6}          & \multicolumn{1}{c|}{70.6}          & \multicolumn{1}{c|}{78.7}          & \multicolumn{1}{c|}{76.1}          & 67.6          & \multicolumn{1}{c|}{76.5}          & \multicolumn{1}{c|}{\textbf{79.9}} & \multicolumn{1}{c|}{86.1}          & \multicolumn{1}{c|}{83.6}          & 72.2          & \multicolumn{1}{c|}{63.4}          & \multicolumn{1}{c|}{45.7}          & \multicolumn{1}{c|}{30.6}          & \multicolumn{1}{c|}{27.6}          & 21.4          \\ \hline
1-Encoded   & \multicolumn{1}{c|}{\textbf{82.7}} & \multicolumn{1}{c|}{\textbf{72.0}} & \multicolumn{1}{c|}{\textbf{79.7}} & \multicolumn{1}{c|}{76.9}          & 67.8          & \multicolumn{1}{c|}{\textbf{77.3}} & \multicolumn{1}{c|}{\textbf{79.9}} & \multicolumn{1}{c|}{\textbf{86.6}} & \multicolumn{1}{c|}{\textbf{84.5}} & \textbf{72.3} & \multicolumn{1}{c|}{64.2}          & \multicolumn{1}{c|}{47.3}          & \multicolumn{1}{c|}{32.4}          & \multicolumn{1}{c|}{29.2}          & 22.4          \\ \hline
4-Encoded & \multicolumn{1}{c|}{82.6}          & \multicolumn{1}{c|}{71.9}          & \multicolumn{1}{c|}{79.5}          & \multicolumn{1}{c|}{\textbf{77.2}} & \textbf{68.6} & \multicolumn{1}{c|}{76.5}          & \multicolumn{1}{c|}{79.7}          & \multicolumn{1}{c|}{85.9}          & \multicolumn{1}{c|}{83.0}          & 71.1          & \multicolumn{1}{c|}{\textbf{66.4}} & \multicolumn{1}{c|}{\textbf{53.0}} & \multicolumn{1}{c|}{\textbf{42.3}} & \multicolumn{1}{c|}{\textbf{38.3}} & \textbf{29.7} \\ \hline
\end{tabular}
\caption{MS-TCN++ results. Improvements in every metric, and a major boost to the Breakfast dataset.}
\label{tab:MS-TCN++ results}
\end{table*}

\begin{table*}[!h]
\centering
\begin{tabular}{|c|ccccc|ccccc|ccccc|}
\hline
          & \multicolumn{5}{c|}{50salads}                                                                                                                                     & \multicolumn{5}{c|}{GTEA}                                                                                                                                         & \multicolumn{5}{c|}{Breakfast}                                                                                                                                    \\ \hline
          & \multicolumn{1}{c|}{Acc}           & \multicolumn{1}{c|}{Edit}          & \multicolumn{3}{c|}{F1@\{10, 25, 50)}                                                   & \multicolumn{1}{c|}{Acc}           & \multicolumn{1}{c|}{Edit}          & \multicolumn{3}{c|}{F1@\{10, 25, 50\}}                                                  & \multicolumn{1}{c|}{Acc}           & \multicolumn{1}{c|}{Edit}          & \multicolumn{3}{c|}{F1@\{10, 25, 50\}}                                                  \\ \hline
Original  & \multicolumn{1}{c|}{\textbf{80.1}} & \multicolumn{1}{c|}{\textbf{71.6}} & \multicolumn{1}{c|}{78.0}          & \multicolumn{1}{c|}{\textbf{75.4}} & \textbf{65.8} & \multicolumn{1}{c|}{77.4}          & \multicolumn{1}{c|}{80.6}          & \multicolumn{1}{c|}{86.7}          & \multicolumn{1}{c|}{\textbf{85.4}} & 73.1          & \multicolumn{1}{c|}{\textbf{68.0}} & \multicolumn{1}{c|}{68.7}          & \multicolumn{1}{c|}{67.5}          & \multicolumn{1}{c|}{60.9}          & 46.3          \\ \hline
1-Encoded   & \multicolumn{1}{c|}{78.9}          & \multicolumn{1}{c|}{70.0}          & \multicolumn{1}{c|}{77.0}          & \multicolumn{1}{c|}{73.7}          & 64.4          & \multicolumn{1}{c|}{76.0}          & \multicolumn{1}{c|}{80.4}          & \multicolumn{1}{c|}{85.7}          & \multicolumn{1}{c|}{84.3}          & 70.2          & \multicolumn{1}{c|}{67.5}          & \multicolumn{1}{c|}{\textbf{69.2}} & \multicolumn{1}{c|}{\textbf{68.5}} & \multicolumn{1}{c|}{\textbf{61.8}} & \textbf{47.0} \\ \hline
4-Encoded & \multicolumn{1}{c|}{79.3}          & \multicolumn{1}{c|}{70.7}          & \multicolumn{1}{c|}{\textbf{78.2}} & \multicolumn{1}{c|}{75.3}          & 65.1          & \multicolumn{1}{c|}{\textbf{77.5}} & \multicolumn{1}{c|}{\textbf{81.6}} & \multicolumn{1}{c|}{\textbf{87.0}} & \multicolumn{1}{c|}{85.1}          & \textbf{74.9} & \multicolumn{1}{c|}{66.5}          & \multicolumn{1}{c|}{66.8}          & \multicolumn{1}{c|}{65.3}          & \multicolumn{1}{c|}{58.4}          & 44.1          \\ \hline
\end{tabular}
\caption{DTGRM results. DTGRM received minor improvements on GTEA and Breakfast.}
\label{tab: DTGRM results}
\end{table*}

Except for DTGRM, which mostly presents modest improvements, most of the highest results are divided between the 1-encoded features and the 4-encoded features, often both are superior the original features. While most of the improvements are of scale 0.5\%-2\%, MS-TCN on Breakfast achieved exceptional results. Both 1-encoded and 4-encoded dramatically increased all metrics. Most importantly, the 4-encoded features has increased the accuracy by 3\%, the edit by 7.3\% and the F1@\{10, 25, 50\} by 11.7\%, 10.7\%, and 8.3\% respectively. These results suggest that we will be able to increase the results on other datasets using other models by searching for better hyper-parameters. We verified this hypothesis in section \ref{sec:hyper}. Moreover, since MS-TCN and MS-TCN++ are used as backbones in many models for many fields, using our methodology on MS-TCN based models might increase the results significantly, regardless of the task and even without hyperparameter search.

\subsubsection{Online Action Segmentation}\label{sec:online}
This section is based on the work of Goldbraikh \etal \cite{twinanda2016endonet,goldbraikh2022bounded}. We reconstructed the paper results on the VTS dataset and compared them to our method. The results Goldbraikh \etal presented show that there is a significant gain from observing a limited window of the future. As of that, we added an experiment of 10-encoded features. Since the original videos of the dataset were captured in 30 frames per second, this is a prediction of 0.33 seconds into the future. The results are presented in table \ref{tab:APAS}.

\begin{table*}[!h]
\begin{tabular}{|c|cccccc||cccccc|}
\hline
                                 & \multicolumn{6}{c||}{Offline}                                                                                                                                                                                 & \multicolumn{6}{c|}{Online}                                                                                                                                                                                  \\ \hline
                                 & \multicolumn{1}{c|}{Acc}            & \multicolumn{1}{c|}{Edit}           & \multicolumn{1}{c|}{F1-macro}       & \multicolumn{3}{c||}{F1@\{10,25,50\}}                                                       & \multicolumn{1}{c|}{Acc}            & \multicolumn{1}{c|}{Edit}           & \multicolumn{1}{c|}{F1-macro}       & \multicolumn{3}{c|}{F1@\{10,25,50\}}                                                       \\ \hline
Original                         & \multicolumn{1}{c|}{86.52}          & \multicolumn{1}{c|}{82.91}          & \multicolumn{1}{c|}{83.48}          & \multicolumn{1}{c|}{87.54}          & \multicolumn{1}{c|}{86.07}          & 79.09          & \multicolumn{1}{c|}{85.03}          & \multicolumn{1}{c|}{63.12}          & \multicolumn{1}{c|}{80.87}          & \multicolumn{1}{c|}{73.24}          & \multicolumn{1}{c|}{71.36}          & 63.92          \\ \hline
1-Encoded                        & \multicolumn{1}{c|}{\textbf{86.87}} & \multicolumn{1}{c|}{\textbf{83.95}} & \multicolumn{1}{c|}{83.55}          & \multicolumn{1}{c|}{\textbf{88.31}} & \multicolumn{1}{c|}{\textbf{86.60}} & 79.09          & \multicolumn{1}{c|}{84.62}          & \multicolumn{1}{c|}{\textbf{64.67}} & \multicolumn{1}{c|}{81.45}          & \multicolumn{1}{c|}{\textbf{74.71}} & \multicolumn{1}{c|}{\textbf{72.39}} & \textbf{64.65} \\ \hline
4-Encoded                        & \multicolumn{1}{c|}{86.14}          & \multicolumn{1}{c|}{83.42}          & \multicolumn{1}{c|}{82.91}          & \multicolumn{1}{c|}{87.63}          & \multicolumn{1}{c|}{85.85}          & 78.70          & \multicolumn{1}{c|}{84.99}          & \multicolumn{1}{c|}{63.11}          & \multicolumn{1}{c|}{81.21}          & \multicolumn{1}{c|}{73.35}          & \multicolumn{1}{c|}{71.22}          & 63.55          \\ \hline
\multicolumn{1}{|l|}{10-Encoded} & \multicolumn{1}{c|}{86.80}          & \multicolumn{1}{c|}{83.42}          & \multicolumn{1}{c|}{\textbf{83.99}} & \multicolumn{1}{c|}{88.07}          & \multicolumn{1}{c|}{86.49}          & \textbf{79.93} & \multicolumn{1}{c|}{\textbf{85.40}} & \multicolumn{1}{c|}{63.56}          & \multicolumn{1}{c|}{\textbf{81.85}} & \multicolumn{1}{c|}{73.66}          & \multicolumn{1}{c|}{71.33}          & 64.47          \\ \hline
\end{tabular}
\caption{Offline and online results on the VTS dataset. Since an online model can gain from future information, we also examined 10-Encoded features, which are predictions of 0.33 seconds into the future.}
\label{tab:APAS}
\end{table*}

The best results are divided between the 1-encoded and the 10-encoded. While the 1-encoded features present up to 1.5\% improvements on all metrics, except accuracy. The 10-encoded features presents an improvement of 1\% in F1-macro, and 0.4\% improvement in accuracy, but exceeded the 1-encoded in those two metrics. 

As there is a strong imbalance between gestures in the VTS dataset, e.g. cutting the suture is a rare class, the F1-macro is most appropriate for evaluating frame-wise performance. That is why 1\% improvement of F1-macro is harder to achieve and is more significant. The model using our features is more suitable for recognizing imbalanced action classes. 


Although the 10-encoded features presents a relatively major improvement in F1-macro, they have a disadvantage. Using the 10th predicted features in real applications may increase runtime. The feature generator can predict up to 300 vectors per second on Nvidia RTX A6000. Because we predicted the 10th next future vector and not the next one, the generator was only able to predict up to 30 vectors per second. This might be problematic for real-time setting, where time is of the essence. The 1-encoded vectors are much faster to create and receive satisfying results, and as of that, they are more recommended for such cases, especially in balanced dataset.

\subsubsection{Timestamp Supervision Action Segmentation}
In this sub-section we present and compare our method's results and the original results using Li \etal. \cite{li2021temporal} work. We tested Li's \etal. model on the 3 cooking domain datasets: GTEA, 50salads and Breakfast.
We tested the 1-encoded and 4-encoded features. The results are presented in tabel \ref{tab:timestamp}.

\begin{table}[]
    \centering
    \begin{tabular}{c|c|c|c|c|c|}
    50salads & Acc & Edit & F1@10 & F1@25 & F1@50 \\
    \hline\hline
    Original & 75.01 & \textbf{67.52} & \textbf{74.88} & \textbf{72.15} & 60.43\\
    1-Encoded & 74.91 & 66.46 & 73.77 & 71.35 & \textbf{60.80} \\
    4-Encoded & \textbf{75.56} & 66.82 & 74.80 & 71.78 & 59.20 \\
    \hline\hline
    GTEA & Acc & Edit & F1@10 & F1@25 & F1@50 \\
    \hline\hline
    Original & 68.12 & 71.26 & 76.07 & 72.50 & 57.21\\
    1-Encoded & 67.86 & 73.70 & 79.11 & 74.78 & 57.78\\
    4-Encoded & \textbf{68.68} & \textbf{74.99} & \textbf{79.53} & \textbf{76.17} & \textbf{60.43}\\
    \hline\hline
    Breakfast & Acc & Edit & F1@10 & F1@25 & F1@50\\
    \hline\hline
    Original & \textbf{64.85} & \textbf{71.21} & \textbf{71.60} &\textbf{ 64.69} & \textbf{48.76} \\
    1-Encoded & 61.31 & 68.74 & 68.56 & 61.61 & 45.67 \\
    4-Encoded & 62.56 & 67.82 & 67.63 & 60.34 & 44.73\\
    \hline

    \end{tabular}
    \caption{Timestamp supervision action segmentation results. Major improvements on GTEA}
    \label{tab:timestamp}
\end{table}
On the one hand, the results on 50salads are ambiguous and on Breakfast there is a drop in performance. On the other hand, on GTEA there is a 0.5\% improvement in accuracy and at least 3\% improvement in the rest of the metrics with the 4-encoded features. For the 1-encoded features, we can also see a performance boost in all metrics, except for accuracy, in the range of 0.5\% - 3\%. However, it is less significant compared to the results achieved with the 4-encoded features.\\
In the next section (\ref{sec:hyper}) we present the results on a hyper parameter tuning experiment. The results on action segmentation with timestamp supervision, using the original hyper parameters, do not show robust improvements (as presented for other sub-tasks in \ref{sec:regular}, \ref{sec:online}). Nevertheless, other results give strong indication for possible improvements for this sub-task using hyperparameter tuning.
For example, on Breakfast Timestamp, which we failed, MS-TCN++ showed large improvements on action segmentation(as presented in \ref{sec:regular}). Moreover, the results presented in the next section (\ref{sec:hyper}) show the benefit of hyperparameter search. Combining these with the results on GTEA suggests that with proper hyperparameter tuning the results can be improved using our method.

\subsubsection{Hyper Parameter Tuning}\label{sec:hyper}
For this section we evaluate the impact of hyper parameter tuning of the temporal model, combined with our new features, on the performance. We experimented on the DTGRM model, which received the most inferior results using our method on 50salads. We attempted to find the optimal parameters for the 50salads dataset. In addition we tested these selected hyperparameters also on the GTEA dataset, which as well received poor results using our method with 1-encoded features. The results are presented in tables \ref{tab:hyper DTGRM 50salads} and \ref{tab:hyper DTGRM gtea}.

\begin{table}[!h]
    \centering
    \begin{tabular}{c|c|c|c|c|c|c}
    Param Set  & Acc & Edit & F1@10 & F1@25 & F1@50 \\
    \hline\hline
    Original & 80.09 & 71.61 & 78.03 & 75.35 & 65.85 \\
    \hline
    New & \textbf{81.42} & \textbf{75.11} & \textbf{81.03} & \textbf{78.39} & \textbf{69.56} \\
    \hline
    \end{tabular}
    \caption{Results on 50salads using the original features and parameters, and the results using the 1-encoded features with the new hyper parameters.}
    \label{tab:hyper DTGRM 50salads}
\end{table}

\begin{table}[!h]
    \centering
    \begin{tabular}{c|c|c|c|c|c|c}
    Param Set & Acc & Edit & F1@10 & F1@25 & F1@50 \\
    \hline\hline
       Original & 77.39 & 80.63 & 86.72 & 85.40 & 73.07\\
       \hline
    New & 76.26 & 83.33 & 88.18 & 86.11 & 73.15 \\
    \hline
    New+10 & \textbf{77.95} & \textbf{85.33} & \textbf{89.20} & \textbf{88.07} & \textbf{77.52}\\
    \hline
    \end{tabular}
    \caption{The results of DTGRM on GTEA using the optimal parameters found for 50salads. New+10 are the results of the new parameter set with an additional 10 epochs.}
    \label{tab:hyper DTGRM gtea}
\end{table}

The original parameters can be found in \cite{wang2021temporal}. The changes to the parameters are as follows: lr of 0.0004, 5 stages, 7 layers, 128 Fmaps, DF-size 3, and 60 epochs.

As shown in table \ref{tab:hyper DTGRM 50salads}, most metrics improved by over 3\% with maximum improvement of over 3.5\%.  Those are immense improvements, particularly when taking into consideration that DTGRM failed on 50salads with our features. Comparing to the results of our 1-encoded features with the original parameters there are improvements of up to 5\%. On GTEA dataset we can see similar results using the parameters found for 50salads, with an increase of up to 2.7\% in some metrics. By allowing the model to train for 20 more epochs (few minutes more) we received a much better improvements with up to 5\% more on Edit, 2.5\% more on F1@10 and F1@25, and 4.4\% more on F1@50. These results suggest that the improvement we achieved using our method without hyper-parameter tuning, are a lower bound of the results that can be achieved by applying our method with proper hyper parameter tuning.

\subsubsection{All Data Self Supervised Training}
 Since future generation is completely self-supervised, the generator can be fine-tuned based on the data we want to classify. We examined the potential advantage of this concept by utilizing the test data as part of the generator training set. Based on that, for each dataset we trained a future generator using both the train and test data, creating a new 1-encoded features set. We named these features 'c-encoded' (combined-encoded). We trained ASFormer, which is to the best of our knowledge the current state-of-the-art model, on those features and compared the results to the original features and the 1-encoded features. The results are presented in table \ref{tab:full}.

\begin{table}[!h]
    \centering
    \begin{tabular}{c|c|c|c|c|c|}
        50salads & Acc & Edit & F1@10 & F1@25 & F1@50 \\
        \hline\hline
        Original & 85.79 & 76.35 & 83.04 & 81.67 & 74.62 \\
        1-Encoded & 85.16 & 78.70 & \textbf{85.01} & \textbf{83.82} & 76.25 \\
        C-Encoded & \textbf{86.64} & \textbf{79.35} & \textbf{85.01} & 83.36 & \textbf{76.53} \\
        \hline\hline
        GTEA & Acc & Edit & F1@10 & F1@25 & F1@50 \\
        \hline\hline
        Original & \textbf{80.15} & 85.58 & 90.75 & 89.82 & 79.48 \\
        1-Encoded & 79.48 & 85.90 & 89.84 & 88.55 & 79.71 \\
        C-Encoded & 79.41 & \textbf{86.74} & \textbf{91.33} & \textbf{89.85} & \textbf{80.93} \\
        \hline\hline
        Breakfast & Acc & Edit & F1@10 & F1@25 & F1@50 \\
        \hline\hline
        Original &  74.01 & 75.45 & 76.64 & 71.66 & 58.92 \\
        1-Encoded & \textbf{74.66} & 75.56 & 77.13 & 71.72 & 59.44 \\
        C-Encoded & 74.01 & \textbf{76.01} & \textbf{77.51} & \textbf{72.22} & \textbf{59.64} \\
        \hline
    \end{tabular}
    \caption{ASFormer results on all 3 datasets. C-Encoded is referring to the features produced from the generator that trained on both train and test sets.}
    \label{tab:full}
\end{table}

Though in most cases, the c-features show relatively minor improvement over the 1-encoded, about 0.5\%-0.9\%, it is mostly consistent and sometimes increases the overall improvement up to 3\%. In addition, these features improved several results where the 1-encoded features failed, for example accuracy on 50salads which increased by approximately 1\%.

\section{Conclusions}
We presented a self-supervised method that comes in the middle of the action segmentation standard pipeline. The generator creates refined representations of the original feature vectors that were used by the temporal model. The experimental evaluations show that our new method is able to enhance the performance of existing temporal models. We show that these improvements are achievable regardless of the temporal model, the sub-task of action segmentation, the domains of the dataset, and the original feature encoders, without additional hyperparameters search.
Furthermore we showed the potential improvement of hyperparameter tuning using our features to all metrics.

{\small
\bibliographystyle{ieee_fullname}
\bibliography{egbib}

\begin{thebibliography}{10}\itemsep=-1pt

\bibitem{bai2018empirical}
Shaojie Bai, J~Zico Kolter, and Vladlen Koltun.
\newblock An empirical evaluation of generic convolutional and recurrent networks for sequence modeling.
\newblock {\em arXiv preprint arXiv:1803.01271}, 2018.

\bibitem{carreira2017quo}
Joao Carreira and Andrew Zisserman.
\newblock Quo vadis, action recognition? a new model and the kinetics dataset.
\newblock In {\em proceedings of the IEEE Conference on Computer Vision and Pattern Recognition}, pages 6299--6308, 2017.

\bibitem{chen2020action}
Min-Hung Chen, Baopu Li, Yingze Bao, Ghassan AlRegib, and Zsolt Kira.
\newblock Action segmentation with joint self-supervised temporal domain adaptation.
\newblock In {\em Proceedings of the IEEE/CVF Conference on Computer Vision and Pattern Recognition}, pages 9454--9463, 2020.

\bibitem{doersch2015unsupervised}
Carl Doersch, Abhinav Gupta, and Alexei~A Efros.
\newblock Unsupervised visual representation learning by context prediction.
\newblock In {\em Proceedings of the IEEE international conference on computer vision}, pages 1422--1430, 2015.

\bibitem{donahue2015long}
Jeffrey Donahue, Lisa Anne~Hendricks, Sergio Guadarrama, Marcus Rohrbach, Subhashini Venugopalan, Kate Saenko, and Trevor Darrell.
\newblock Long-term recurrent convolutional networks for visual recognition and description.
\newblock In {\em Proceedings of the IEEE conference on computer vision and pattern recognition}, pages 2625--2634, 2015.

\bibitem{farha2019ms}
Yazan~Abu Farha and Jurgen Gall.
\newblock Ms-tcn: Multi-stage temporal convolutional network for action segmentation.
\newblock In {\em Proceedings of the IEEE/CVF Conference on Computer Vision and Pattern Recognition}, pages 3575--3584, 2019.

\bibitem{fathi2011learning}
Alireza Fathi, Xiaofeng Ren, and James~M Rehg.
\newblock Learning to recognize objects in egocentric activities.
\newblock In {\em CVPR 2011}, pages 3281--3288. IEEE, 2011.

\bibitem{finn2016unsupervised}
Chelsea Finn, Ian Goodfellow, and Sergey Levine.
\newblock Unsupervised learning for physical interaction through video prediction.
\newblock {\em Advances in neural information processing systems}, 29, 2016.

\bibitem{gao2021future}
Xiaojie Gao, Yueming Jin, Zixu Zhao, Qi Dou, and Pheng-Ann Heng.
\newblock Future frame prediction for robot-assisted surgery.
\newblock In {\em International Conference on Information Processing in Medical Imaging}, pages 533--544. Springer, 2021.

\bibitem{gao2014jhu}
Yixin Gao, S~Swaroop Vedula, Carol~E Reiley, Narges Ahmidi, Balakrishnan Varadarajan, Henry~C Lin, Lingling Tao, Luca Zappella, Benjam{\i}n B{\'e}jar, David~D Yuh, et~al.
\newblock Jhu-isi gesture and skill assessment working set (jigsaws): A surgical activity dataset for human motion modeling.
\newblock In {\em MICCAI workshop: M2cai}, volume~3, 2014.

\bibitem{gidaris2018unsupervised}
Spyros Gidaris, Praveer Singh, and Nikos Komodakis.
\newblock Unsupervised representation learning by predicting image rotations.
\newblock In {\em ICLR 2018}, 2018.

\bibitem{girdhar2021anticipative}
Rohit Girdhar and Kristen Grauman.
\newblock Anticipative video transformer.
\newblock In {\em Proceedings of the IEEE/CVF International Conference on Computer Vision}, pages 13505--13515, 2021.

\bibitem{goldbraikh2022bounded}
Adam Goldbraikh, Netanell Avisdris, Carla~M Pugh, and Shlomi Laufer.
\newblock Bounded future ms-tcn++ for surgical gesture recognition.
\newblock {\em arXiv preprint arXiv:2209.14647}, 2022.

\bibitem{goldbraikh2022video}
Adam Goldbraikh, Anne-Lise D’Angelo, Carla~M Pugh, and Shlomi Laufer.
\newblock Video-based fully automatic assessment of open surgery suturing skills.
\newblock {\em International Journal of Computer Assisted Radiology and Surgery}, 17(3):437--448, 2022.

\bibitem{hosseini2019inception}
Matin Hosseini, Anthony~S Maida, Majid Hosseini, and Gottumukkala Raju.
\newblock Inception-inspired lstm for next-frame video prediction.
\newblock {\em arXiv preprint arXiv:1909.05622}, 2019.

\bibitem{huynh2008scope}
Quan Huynh-Thu and Mohammed Ghanbari.
\newblock Scope of validity of psnr in image/video quality assessment.
\newblock {\em Electronics letters}, 44(13):800--801, 2008.

\bibitem{karaman2014fast}
Svebor Karaman, Lorenzo Seidenari, and Alberto Del~Bimbo.
\newblock Fast saliency based pooling of fisher encoded dense trajectories.
\newblock In {\em ECCV THUMOS Workshop}, volume~1, page~5, 2014.

\bibitem{ke2019time}
Qiuhong Ke, Mario Fritz, and Bernt Schiele.
\newblock Time-conditioned action anticipation in one shot.
\newblock In {\em Proceedings of the IEEE/CVF Conference on Computer Vision and Pattern Recognition}, pages 9925--9934, 2019.

\bibitem{kolesnikov2019revisiting}
Alexander Kolesnikov, Xiaohua Zhai, and Lucas Beyer.
\newblock Revisiting self-supervised visual representation learning.
\newblock In {\em Proceedings of the IEEE/CVF conference on computer vision and pattern recognition}, pages 1920--1929, 2019.

\bibitem{kuehne2014language}
Hilde Kuehne, Ali Arslan, and Thomas Serre.
\newblock The language of actions: Recovering the syntax and semantics of goal-directed human activities.
\newblock In {\em Proceedings of the IEEE conference on computer vision and pattern recognition}, pages 780--787, 2014.

\bibitem{kwon2019predicting}
Yong-Hoon Kwon and Min-Gyu Park.
\newblock Predicting future frames using retrospective cycle gan.
\newblock In {\em Proceedings of the IEEE/CVF Conference on Computer Vision and Pattern Recognition}, pages 1811--1820, 2019.

\bibitem{lea2017temporal}
Colin Lea, Michael~D Flynn, Rene Vidal, Austin Reiter, and Gregory~D Hager.
\newblock Temporal convolutional networks for action segmentation and detection.
\newblock In {\em proceedings of the IEEE Conference on Computer Vision and Pattern Recognition}, pages 156--165, 2017.

\bibitem{li2020ms}
Shi-Jie Li, Yazan AbuFarha, Yun Liu, Ming-Ming Cheng, and Juergen Gall.
\newblock Ms-tcn++: Multi-stage temporal convolutional network for action segmentation.
\newblock {\em IEEE Transactions on Pattern Analysis and Machine Intelligence}, pages 1--1, 2020.

\bibitem{li2021temporal}
Zhe Li, Yazan Abu~Farha, and Jurgen Gall.
\newblock Temporal action segmentation from timestamp supervision.
\newblock In {\em Proceedings of the IEEE/CVF Conference on Computer Vision and Pattern Recognition}, pages 8365--8374, 2021.

\bibitem{liang2017dual}
Xiaodan Liang, Lisa Lee, Wei Dai, and Eric~P Xing.
\newblock Dual motion gan for future-flow embedded video prediction.
\newblock In {\em proceedings of the IEEE international conference on computer vision}, pages 1744--1752, 2017.

\bibitem{lim2021time}
Bryan Lim and Stefan Zohren.
\newblock Time-series forecasting with deep learning: a survey.
\newblock {\em Philosophical Transactions of the Royal Society A}, 379(2194):20200209, 2021.

\bibitem{lotter2017deep}
William Lotter, Gabriel Kreiman, and David Cox.
\newblock Deep predictive coding networks for video prediction and unsupervised learning.
\newblock In {\em International Conference on Learning Representations}, 2017.

\bibitem{ma2021towards}
Pingchuan Ma, Brais Martinez, Stavros Petridis, and Maja Pantic.
\newblock Towards practical lipreading with distilled and efficient models.
\newblock In {\em ICASSP 2021-2021 IEEE International Conference on Acoustics, Speech and Signal Processing (ICASSP)}, pages 7608--7612. IEEE, 2021.

\bibitem{mathieu2016deep}
Michael Mathieu, Camille Couprie, and Yann LeCun.
\newblock Deep multi-scale video prediction beyond mean square error.
\newblock In {\em 4th International Conference on Learning Representations, ICLR 2016}, 2016.

\bibitem{micikevicius2017mixed}
Paulius Micikevicius, Sharan Narang, Jonah Alben, Gregory Diamos, Erich Elsen, David Garcia, Boris Ginsburg, Michael Houston, Oleksii Kuchaiev, Ganesh Venkatesh, et~al.
\newblock Mixed precision training.
\newblock {\em arXiv preprint arXiv:1710.03740}, 2017.

\bibitem{moltisanti2019action}
Davide Moltisanti, Sanja Fidler, and Dima Damen.
\newblock Action recognition from single timestamp supervision in untrimmed videos.
\newblock In {\em Proceedings of the IEEE/CVF Conference on Computer Vision and Pattern Recognition}, pages 9915--9924, 2019.

\bibitem{rohrbach2012database}
Marcus Rohrbach, Sikandar Amin, Mykhaylo Andriluka, and Bernt Schiele.
\newblock A database for fine grained activity detection of cooking activities.
\newblock In {\em 2012 IEEE conference on computer vision and pattern recognition}, pages 1194--1201. IEEE, 2012.

\bibitem{singh2016multi}
Bharat Singh, Tim~K Marks, Michael Jones, Oncel Tuzel, and Ming Shao.
\newblock A multi-stream bi-directional recurrent neural network for fine-grained action detection.
\newblock In {\em Proceedings of the IEEE conference on computer vision and pattern recognition}, pages 1961--1970, 2016.

\bibitem{srivastava2015unsupervised}
Nitish Srivastava, Elman Mansimov, and Ruslan Salakhudinov.
\newblock Unsupervised learning of video representations using lstms.
\newblock In {\em International conference on machine learning}, pages 843--852. PMLR, 2015.

\bibitem{stein2013combining}
Sebastian Stein and Stephen~J McKenna.
\newblock Combining embedded accelerometers with computer vision for recognizing food preparation activities.
\newblock In {\em Proceedings of the 2013 ACM international joint conference on Pervasive and ubiquitous computing}, pages 729--738, 2013.

\bibitem{twinanda2016endonet}
Andru~P Twinanda, Sherif Shehata, Didier Mutter, Jacques Marescaux, Michel De~Mathelin, and Nicolas Padoy.
\newblock Endonet: a deep architecture for recognition tasks on laparoscopic videos.
\newblock {\em IEEE transactions on medical imaging}, 36(1):86--97, 2016.

\bibitem{villegas2017learning}
Ruben Villegas, Jimei Yang, Yuliang Zou, Sungryull Sohn, Xunyu Lin, and Honglak Lee.
\newblock Learning to generate long-term future via hierarchical prediction.
\newblock In {\em international conference on machine learning}, pages 3560--3569. PMLR, 2017.

\bibitem{vinyals2015show}
Oriol Vinyals, Alexander Toshev, Samy Bengio, and Dumitru Erhan.
\newblock Show and tell: A neural image caption generator.
\newblock In {\em Proceedings of the IEEE conference on computer vision and pattern recognition}, pages 3156--3164, 2015.

\bibitem{vondrick2016generating}
Carl Vondrick, Hamed Pirsiavash, and Antonio Torralba.
\newblock Generating videos with scene dynamics.
\newblock {\em Advances in neural information processing systems}, 29, 2016.

\bibitem{wang2021temporal}
Dong Wang, Di Hu, Xingjian Li, and Dejing Dou.
\newblock Temporal relational modeling with self-supervision for action segmentation.
\newblock In {\em Proceedings of the AAAI Conference on Artificial Intelligence}, volume~35, pages 2729--2737, 2021.

\bibitem{wang2004image}
Zhou Wang, Alan~C Bovik, Hamid~R Sheikh, and Eero~P Simoncelli.
\newblock Image quality assessment: from error visibility to structural similarity.
\newblock {\em IEEE transactions on image processing}, 13(4):600--612, 2004.

\bibitem{wang2020boundary}
Zhenzhi Wang, Ziteng Gao, Limin Wang, Zhifeng Li, and Gangshan Wu.
\newblock Boundary-aware cascade networks for temporal action segmentation.
\newblock In {\em European Conference on Computer Vision}, pages 34--51. Springer, 2020.

\bibitem{xu2019self}
Dejing Xu, Jun Xiao, Zhou Zhao, Jian Shao, Di Xie, and Yueting Zhuang.
\newblock Self-supervised spatiotemporal learning via video clip order prediction.
\newblock In {\em Proceedings of the IEEE/CVF Conference on Computer Vision and Pattern Recognition}, pages 10334--10343, 2019.

\bibitem{chinayi_ASformer}
Fangqiu Yi, Hongyu Wen, and Tingting Jiang.
\newblock Asformer: Transformer for action segmentation.
\newblock In {\em The British Machine Vision Conference (BMVC)}, 2021.

\bibitem{zbontar2021barlow}
Jure Zbontar, Li Jing, Ishan Misra, Yann LeCun, and St{\'e}phane Deny.
\newblock Barlow twins: Self-supervised learning via redundancy reduction.
\newblock In {\em International Conference on Machine Learning}, pages 12310--12320. PMLR, 2021.

\bibitem{zhang2021ms}
Jiyang Zhang, Yuxuan Wang, Jianxiong Tang, Jianxiao Zou, and Shicai Fan.
\newblock Ms-tcn: A multiscale temporal convolutional network for fault diagnosis in industrial processes.
\newblock In {\em 2021 American Control Conference (ACC)}, pages 1601--1606. IEEE, 2021.

\end{thebibliography}
}

\end{document}